\documentclass{article}

\usepackage[main,final]{neurips_2026}

\usepackage[utf8]{inputenc} 
\usepackage[T1]{fontenc}    
\usepackage{xurl}           
\usepackage{courier}        
\usepackage{booktabs}       
\usepackage{amsfonts}       
\usepackage{nicefrac}       
\usepackage{microtype}      
\usepackage[table]{xcolor}  
\usepackage{amsmath,amssymb,bm}
\usepackage{amsthm}
\usepackage{graphicx}

\usepackage{multirow}        
\usepackage{pifont}
                   
\newcommand{\xmark}{\ding{55}}
\usepackage{array} 
\usepackage{subcaption}

\usepackage{fontawesome5}
\usepackage{algorithm}
\usepackage{algpseudocode}
\usepackage[hidelinks]{hyperref}
\hypersetup{
  pdftitle={Parallel Rollout Approximation for Pixel-Space Autoregressive Image Generation},
  pdfauthor={Jiayi Xu, Di He, Guolin Ke},
  pdfsubject={NeurIPS 2026 camera-ready paper},
  pdfkeywords={autoregressive image generation, pixel space, parallel rollout approximation}
}
\newcommand{\scitep}[1]{{\scriptsize\citep{#1}}}

\makeatletter\def\shline{\noalign{\ifnum0=`}\fi\hrule\@height 1pt \futurelet\reserved@a\@xhline}\def\midline{\noalign{\ifnum0=`}\fi\hrule \@height 0.4pt \futurelet\reserved@a\@xhline}\makeatother

\title{Parallel Rollout Approximation for Pixel-Space Autoregressive Image Generation}

\author{%
  Jiayi Xu$^{1,2}$ \quad
  Di He$^{1}$ \quad
  \textbf{Guolin Ke}$^{2}$\thanks{\parbox[t]{0.85\linewidth}{%
        Corresponding author. Email: \texttt{kegl@dp.tech}\\
         \faGithub\ \textbf{Code:} \url{https://github.com/MangataX/PRA}%
      }%
 }\\
  \quad $^1$ Peking University \quad
  \quad $^2$ DP Technology 
  \\
  }

\begin{document}

\maketitle

\begin{abstract}
Pixel-space continuous-token autoregressive (AR) generation directly models images as sequences of raw pixel patches, avoiding discrete tokenization or a separately pretrained tokenizer. However, it faces coupled challenges: high-dimensional patch generation causes large single-step errors, and teacher-forced training creates a train--inference gap that makes these errors accumulate across AR steps. Existing fixes such as $x$-prediction and input noise injection only partially mitigate these issues. Exact rollout training better matches inference-time conditions, but is impractical due to prohibitively slow sequential sampling.
We propose \emph{Parallel Rollout Approximation} (PRA), a scalable framework that addresses both challenges jointly. PRA generates low-dimensional intermediate states and maps them to pixel patches through a one-step denoising pixel decoder, preserving a pixel-in, pixel-out AR interface. The same decoder also constructs inference-like training inputs from perturbed intermediate states through the same intermediate-state-to-pixel path used at inference, while retaining parallel training over positions. On class-conditional ImageNet-1K generation at $256\times256$ resolution, PRA-S with 135M parameters achieves an FID of 2.58, surpassing the previous billion-scale pixel-space AR result of 3.60. Scaling to PRA-L with 511M parameters further improves FID to 1.94, establishing a new SOTA among pixel-space AR models. Beyond generation, PRA achieves higher ImageNet classification probing accuracy than other AR and diffusion baselines, suggesting its potential for unified pixel-space image generation and understanding.
\end{abstract}

\vspace{-4pt}
\begin{figure}[h]
  \centering
  \includegraphics[width=0.82\linewidth]{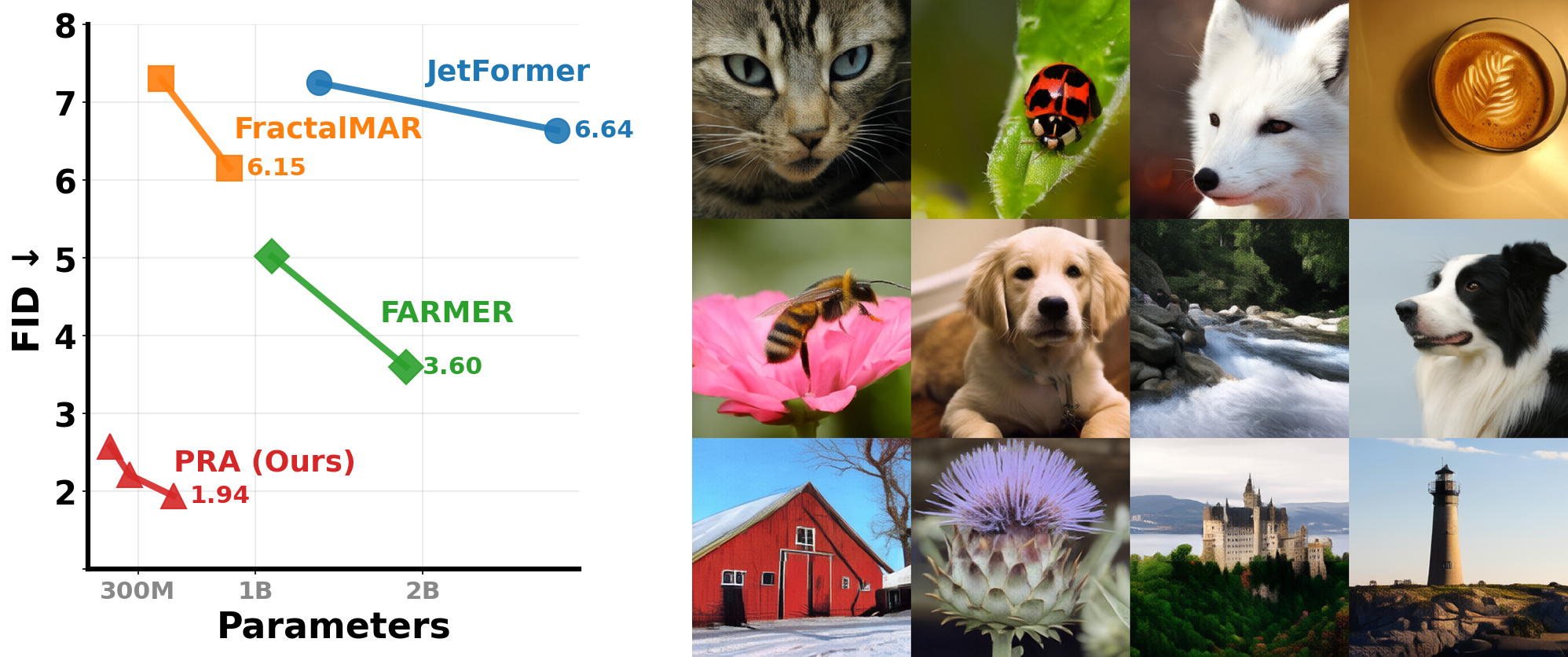}
  \vspace{-2pt}
    \caption{
    \textbf{Left:} FID comparison of pixel-space AR models across parameter scales. PRA achieves substantially lower FID than prior baselines, with PRA-S (135M) already outperforming billion-parameter models.
    \textbf{Right:} Uncurated 256$\times$256 samples generated by PRA-L.
    }
\label{fig:teaser}
\end{figure}
    
\section{Introduction}
\vspace{-10pt}

The success of autoregressive (AR) generation in large language models (LLMs) \citep{radford2018improving,brown2020language} has motivated extending this paradigm to image generation and other continuous domains \citep{li2024autoregressive,meng2024autoregressive,teng2025magi,lu2025uni}. Most successful AR image generators, however, operate in a discrete or latent token space rather than directly in pixel space. Discrete-token approaches rely on vector quantization or pretrained tokenizers to convert images into codebook indices \citep{gray1984vector,van2017neural,esser2021taming,yu2021vector}, while recent continuous-token AR methods typically model continuous tokens in learned latent or feature spaces \citep{tschannen2024givt,li2024autoregressive,sun2024multimodal}. Although these token spaces make autoregressive modeling more tractable, they introduce an additional stage and make the final generation quality constrained by the tokenizer or autoencoder. This raises a natural question: can we build a competitive pixel-space autoregressive model trained end-to-end on raw pixel patches, without relying on an external pretrained tokenizer?
\looseness=-1

\begin{figure}[t]
  \centering
  \includegraphics[width=1.0\linewidth]{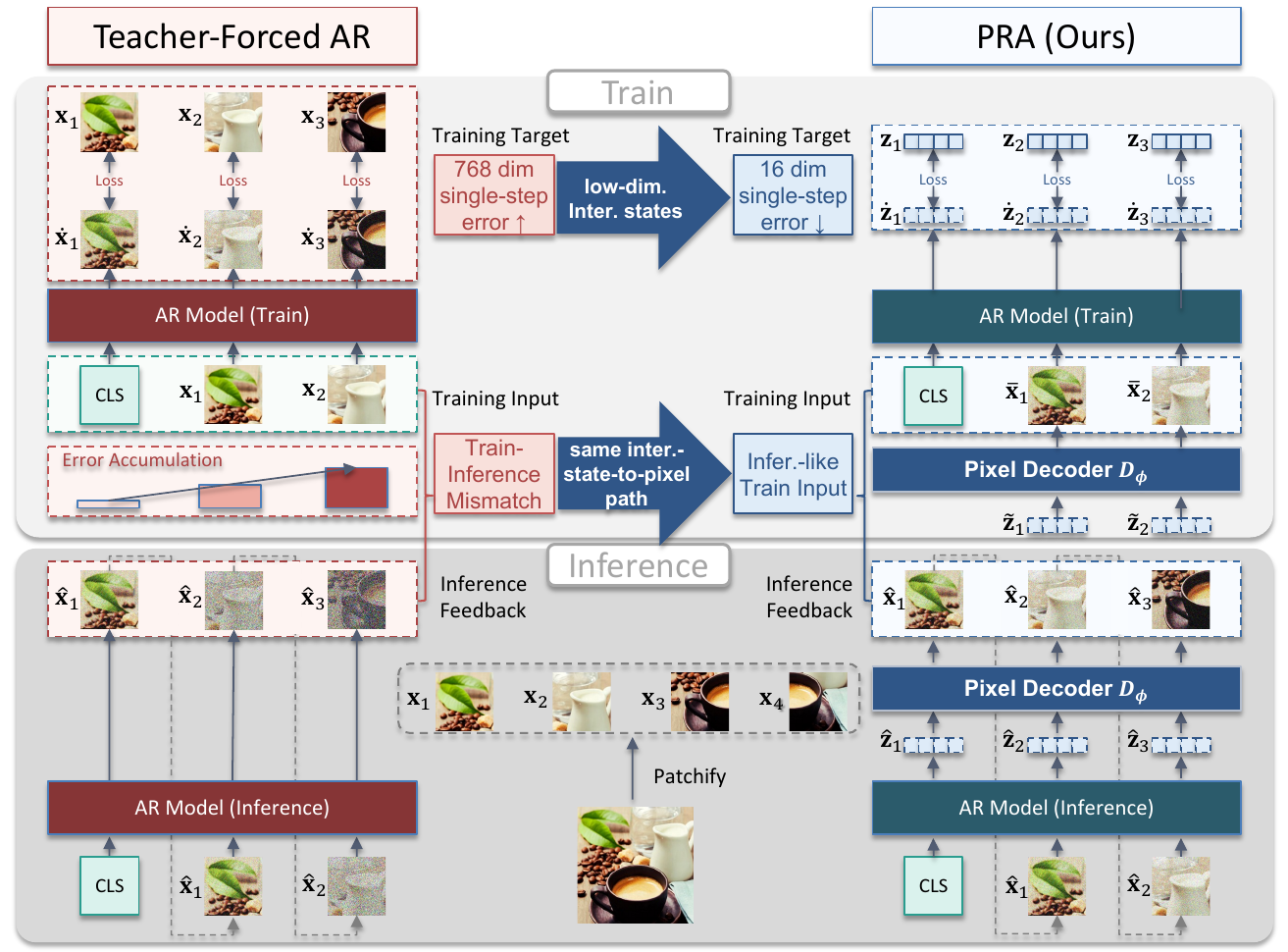}
  \vspace{-10pt}
\caption{
PRA addresses coupled challenges in pixel-space AR generation. \textbf{Output-side:} Low-dimensional intermediate states are generated and decoded into pixel patches. \textbf{Input-side:} Perturbed states pass through the same pixel decoder used at inference to construct inference-like pixel prefixes in parallel. This preserves the pixel-in, pixel-out AR interface without sequential training rollouts.
}
\label{fig:main}
\end{figure}

Pixel-space autoregressive generation appears conceptually simple: an image can be divided into patches, and the model predicts the next raw pixel patch conditioned on previous ones. In this process, raw pixel patches play a dual role: they are the continuous targets to be generated at the current step, and once generated, they become part of the causal context for future steps. As illustrated in Figure~\ref{fig:main}, this dual role exposes two coupled challenges. On the output-side, each raw pixel patch is a high-dimensional continuous vector, which is hard to predict in a single AR step and causes large single-step errors. On the input-side, teacher-forced training conditions the AR model on clean ground-truth prefixes, whereas inference requires conditioning  on previously generated pixel patches, creating a train--inference mismatch. Together, these challenges form a self-reinforcing mode: difficult high-dimensional predictions introduce large local errors, and the autoregressive loop turns these errors into imperfect future contexts, allowing them to propagate and compound across subsequent steps~\citep{bengio2015scheduled,ranzato2016sequence}.
\looseness=-1

 Existing techniques improve pixel-space AR baselines but only partially resolve this coupled difficulty. Prediction parameterizations such as $x$-prediction with the $v$-loss, which are effective in pixel-space diffusion models~\citep{li2025back}, reduce high-dimensional output error but still leave a large gap to diffusion baselines. For the input-side mismatch, injecting noise into ground-truth tokens exposes the AR model to imperfect contexts during training~\citep{bengio2015scheduled, ke2026hyperspherical,pasini2024continuousautoregressivemodelsnoise}, but these perturbations are independent of the model's own generation process and therefore cannot fully match the structured, model-dependent errors encountered during inference-time rollout. Naive on-policy rollout would directly expose the model to the inference-time generated prefixes, but it is impractical for continuous-token AR as it requires prohibitively slow sequential autoregressive sampling with multi-step generation \citep{li2024autoregressive} for each token. 
\looseness=-1

We propose Parallel Rollout Approximation (PRA), a scalable framework for pixel-space autoregressive generation that addresses both bottlenecks jointly. As shown in Figure~\ref{fig:main}, PRA generates compact intermediate states and maps them to pixel patches through a one-step denoising pixel decoder, reducing the difficulty of each autoregressive prediction. The key design is to reuse the pixel decoder to construct training prefixes from perturbed intermediate states, so the AR model learns from pixel contexts produced through the same intermediate-state-to-pixel path as its inference-time feedback. This design addresses prediction difficulty and feedback mismatch together, while preserving parallel training and a pixel-in, pixel-out interface.
\looseness=-1

We validate PRA on class-conditional ImageNet-1K generation at $256\!\times\!256$ resolution, using raw pixel patches as continuous tokens \citep{deng2009imagenet}. PRA substantially improves the quality of pixel-space AR generation across model scales: the 135M-parameter PRA-S already surpasses prior billion-scale pixel-space AR models, and PRA-L reaches an FID of 1.94. Moreover, ablations show that low-dimensional prediction targets alone are insufficient in the pixel-feedback setting, and that constructing training inputs through the shared intermediate-state-to-pixel path is important for effective generation. The generation-trained backbone also achieves stronger ImageNet linear probing accuracy than AR and diffusion baselines\citep{ke2026hyperspherical,li2025back,Peebles_2023_ICCV}, suggesting the promise of end-to-end pixel-space autoregressive learning for both generation and visual understanding.
\looseness=-1

\section{Related Work}
\vspace{-9pt}

\paragraph{Train--Inference Mismatch and Rollout-Based Training.}
Teacher-forced autoregressive training creates a mismatch between training and inference: models are trained on ground-truth prefixes but must condition on their own generated prefixes at inference time, leading to exposure bias and error accumulation~\citep{bengio2015scheduled,ranzato2016sequence,schmidt2019generalization}. Existing remedies expose models to generated or perturbed inputs during training~\citep{he2025rear,ren2025beyond}, including scheduled sampling, sequence-level rollout training, adversarial dynamics matching, and recent on-policy distillation~\citep{bengio2015scheduled,ranzato2016sequence,goyal2016professor,agarwal2024onpolicy}. However, exact rollout-based training is inherently sequential and becomes especially expensive for continuous-token AR models, where each token may require multi-step diffusion sampling. In contrast, PRA constructs inference-like pixel inputs in parallel through the same pixel decoder used for inference-time feedback, avoiding sequential autoregressive rollouts during training.

\vspace{-9pt}
\paragraph{Continuous-Token Autoregressive Generation.}
Continuous-token AR models extend autoregressive generation from discrete vocabularies to continuous-valued tokens, such as latent features or raw signal patches. Recent methods commonly use token-level diffusion heads to model continuous next-token distributions~\citep{li2024autoregressive}. Many strong continuous-token AR methods operate in latent space: they rely on a pretrained tokenizer to map raw signals into lower-dimensional continuous tokens, and then train the AR model on these latent sequences. Prior work further shows that such latent spaces must be carefully designed for stable AR generation, using techniques such as stronger noise injection~\citep{sun2024multimodal,pasini2024continuousautoregressivemodelsnoise} or constant-norm latent representations~\citep{ke2026hyperspherical,team2025nextstep}. These designs improve robustness by making continuous-token generation easier or less sensitive to accumulated errors. PRA instead targets pixel-space continuous-token AR: it learns low-dimensional intermediate states end-to-end with the AR model, without a separately pretrained tokenizer. These states serve as internal prediction targets, while decoded pixels are fed back to the AR backbone, preserving a pixel-in, pixel-out interface.

\vspace{-9pt}
\paragraph{Pixel-Space Generative Modeling.}
Pixel-space image generation has been dominated by diffusion models, which generate images through iterative denoising directly in the original pixel space~\citep{li2025back,wang2025pixnerd,chen2025pixelflow}. Pixel-space diffusion models also handle high-dimensional pixel outputs, but they do not feed locally generated patches back into a causal context across spatial positions. Pixel-space autoregressive generation, in contrast, remains relatively underexplored and has generally underperformed diffusion-based models~\citep{tschannen2025jetformerautoregressivegenerativemodel,li2025fractal,zheng2025farmer}. A key bottleneck is that each generated high-dimensional patch becomes part of the context for later predictions, making AR models vulnerable to cross-step error accumulation. PRA improves the viability of pixel-space AR by reducing per-step generation difficulty with low-dimensional intermediate states and by constructing inference-like pixel inputs in parallel to mitigate train--inference mismatch.

\section{Pixel-Space Image Generation via Continuous-Token AR Modeling}
\vspace{-9pt}

Rather than generating images inefficiently at the pixel level, we adopt a patch-wise formulation \citep{li2025back}, where all pixels within each patch are grouped into a continuous token. For example, with a patch size of $16^2$, each token corresponds to a $16^2 \times 3 = 768$-dim. continuous vector. Given an image, we arrange its patch-wise continuous tokens in raster-scan order as
$\mathbf{x}=(\mathbf{x}_1,\ldots,\mathbf{x}_T)$ with $\mathbf{x}_i \in \mathbb{R}^{d}$. Under the formulation, pixel-space AR image generation becomes a continuous-token autoregressive generation problem, with the joint distribution factorized as
$p(\mathbf{x}) = \prod_{i=1}^{T} p(\mathbf{x}_i \mid \mathbf{x}_{<i})$. \looseness=-1

We use $\mathbf{x}_i$ to denote a ground-truth continuous token and $\hat{\mathbf{x}}_i$ to denote a model-generated continuous token. A causal Transformer encodes the prefix tokens. Under teacher forcing, the model takes the ground-truth prefix $\mathbf{x}_{<i}$ as input and produces a hidden state $\mathbf{h}_{i-1}=f_\theta(\mathbf{x}_{<i})$, where $\theta$ denotes the Transformer parameters. Optional conditioning information, such as class labels or text prompts, can be prepended to the token sequence and included in the causal context. \looseness=-1

To model the continuous next-token distribution, we attach a token-level diffusion head to the AR Transformer \citep{li2024autoregressive}. Conditioned on $\mathbf{h}_{i-1}$, the head learns to transform a simple prior sample into the target token $\mathbf{x}_i$. We train this head with a rectified-flow objective~\citep{liu2022flow}. Given a prior sample $\mathbf{x}_i^0 \sim \mathcal{N}(\mathbf{0},\mathbf{I})$, the target token $\mathbf{x}_i^1=\mathbf{x}_i$, and a time variable $t$, we form the interpolated state $\mathbf{x}_i^t=(1-t)\mathbf{x}_i^0+t\mathbf{x}_i^1$. The diffusion head $v_\omega$ takes $(\mathbf{x}_i^t,t,\mathbf{h}_{i-1})$ as input and is trained to predict either the straight-path velocity $\mathbf{x}_i^1-\mathbf{x}_i^0$ or the clean target $\mathbf{x}_i^1$.

At inference time, the model conditions on previously generated tokens $\hat{\mathbf{x}}_{<i}$ and computes $\hat{\mathbf{h}}_{i-1}=f_\theta(\hat{\mathbf{x}}_{<i})$. The diffusion head samples the next token from Gaussian noise using an Euler--Maruyama sampler followed by a final deterministic Euler step (Appendix~\ref{app:sampling_details}). The final state is taken as the generated continuous token $\hat{\mathbf{x}}_i$, which is then fed back as input for subsequent autoregressive steps. This differs from teacher-forced training, where the model conditions on ground-truth prefixes $\mathbf{x}_{<i}$, creating a train--inference gap that can cause errors to accumulate during generation.

\section{Challenges of Pixel-Space Autoregressive Generation}
\label{sec:diag}
\vspace{-9pt}

Although the continuous-token AR formulation is conceptually straightforward, we find that naive pixel-space AR degrades substantially at higher resolutions. To systematically diagnose this degradation, we compare against JiT \citep{li2025back}, a pixel-space diffusion model, under two controlled settings: $64^2$ resolution with a $4^2$ patch size, and $256^2$ resolution with a $16^2$ patch size. These settings have the same autoregressive length, $(64/4)^2=(256/16)^2=256$, but very different token dimensionalities, $4^2 \times 3 = 48$ versus $16^2 \times 3 = 768$. For fairness, the AR model and JiT use the same model scale (130M parameters) and the same training budget (200 epochs) on ImageNet \citep{deng2009imagenet}. We report FID computed on 50k generated samples, with classifier-free guidance scales tuned for each model. We then introduce diagnostic variants to isolate the two challenges: output-side token dimensionality and input-side train--inference mismatch, with results summarized in Table~\ref{tab:pixel_ar_diagnostics}. \looseness=-1

\vspace{-6pt}
\paragraph{Output-side challenge: high-dimensional tokens increase single-step errors.}
As shown in Table~\ref{tab:pixel_ar_diagnostics}, the AR model is competitive with JiT when the token dimensionality is low, but falls much further behind when the token dimensionality increases from $48$ to $768$. Since both settings use the same number of autoregressive steps, this degradation cannot be explained by a longer generation horizon. Instead, it points to high-dimensional continuous-token generation as a key output-side bottleneck for pixel-space AR.
We further compare the high-dimensional AR setting with and without the $x$-prediction variant from JiT. Similar to its effect in diffusion models, $x$-prediction improves AR performance from 9.70 to 7.68 compared with $v$-prediction (A3 vs. A4). However, the gap to JiT remains large, suggesting that this output-side fix alone is insufficient: directly generating high-dimensional pixel tokens still causes substantial single-step errors.

\vspace{-6pt}
\paragraph{Input-side challenge: train--inference mismatch amplifies errors across steps.}
The input-side difficulty comes from the mismatch between training on ground-truth prefixes and inference on generated prefixes. Once a generated token is fed back into the autoregressive context, its error can influence later predictions and accumulate across steps. We therefore strengthen the AR baseline with input noise injection: each ground-truth input token is perturbed along the rectified-flow interpolation used by the prediction head, $\tilde{\mathbf{x}}_i=(1-t_i)\mathbf{x}_i^0+t_i\mathbf{x}_i$, where $\mathbf{x}_i^0\sim\mathcal{N}(\mathbf{0},\mathbf{I})$ and $t_i\sim\mathcal{U}(t_{\min},1)$, before being fed into the causal context. This improves FID from 9.94 to 7.68 (A6 to A3), but the gap to JiT remains substantial, indicating that independent noise perturbations cannot fully match the structured, model-dependent errors produced during inference-time rollout.

These diagnostics show that pixel-space AR requires addressing both sides of the problem: reducing the difficulty of generating high-dimensional pixel tokens, and training the model under inputs that better resemble the generated tokens it receives during inference-time rollout. PRA tackles these two issues jointly by generating compact intermediate states and constructing decoded inference-like inputs in parallel.

\vspace{-6pt}

\begin{table}[t]
    \centering
    \small
    \caption{
    Diagnostics of pixel-space AR generation.
    All settings use 256 autoregressive tokens.
    Res/Patch denotes resolution/patch size, Noise denotes input noise injection, Tok. Dim. denotes token dimensionality.
    Output-side diagnostics compare JiT and AR under different token dimensionalities.
    Input-side diagnostics compare AR runs with and without input noise injection.
    }
    \label{tab:pixel_ar_diagnostics}
    \setlength{\tabcolsep}{2.8pt}
    \begin{tabular}{llccccccc}
    \toprule
    Study & ID & Res/Patch & Model & Pred. & Noise & Tok. Dim. & Epoch & FID $\downarrow$  \\
    \midrule
    \multirow{6}{*}{Output}
    & D1 & 64/4   & Diffusion (JiT) & $x$ & --          & 48  & 200 & 3.55 \\
    & A1 & 64/4   & AR  & $x$ & \checkmark & 48  & 200 & 4.06 \\
    & A2 & 64/4   & AR  & $v$ & \checkmark & 48  & 200 & 4.15  \\
    \cmidrule(lr){2-9}
    & D2 & 256/16 & Diffusion (JiT) & $x$ & --          & 768 & 200 & 4.56 \\
    & A3 & 256/16 & AR  & $x$ & \checkmark & 768 & 200 & 7.68 \\
    & A4 & 256/16 & AR  & $v$ & \checkmark & 768 & 200 & 9.70 \\
    \midrule
    \multirow{2}{*}{Input}
    & A5 & 64/4   & AR  & $x$ & \xmark     & 48  & 200 & 6.71 \\
    & A6 & 256/16 & AR  & $x$ & \xmark     & 768 & 200 & 9.94 \\
    \bottomrule
    \end{tabular}
\end{table}

\section{Parallel Rollout Approximation}
\label{sec:pra}

Sec.~\ref{sec:diag} identifies two coupled challenges: high-dimensional patch generation and train--inference mismatch. PRA addresses the output-side challenge by generating low-dimensional intermediate states and decoding them into pixel patches. To address the input-side challenge, the same causal pixel decoder constructs inference-like training inputs in parallel, avoiding sequential rollouts while preserving the pixel-in, pixel-out AR interface.

\subsection{End-to-End Intermediate Targets for Pixel Outputs}
\label{sec:pra_low_dim}

PRA first addresses the output-side difficulty of directly generating high-dimensional pixel patches by introducing an end-to-end learned intermediate target. As illustrated in Figure~\ref{fig:main}, for each ground-truth pixel token $\mathbf{x}_i\in\mathbb{R}^{d}$, we construct a low-dimensional state $\mathbf{z}_i\in\mathbb{R}^{d_z}$ with $d_z<d$, and use a causal pixel decoder $D_\phi$ to map it back to the pixel-token space. The decoder is causal and trained to reconstruct the original token, i.e., $\mathbf{x}^{\mathrm{rec}}_i=D_\phi(\mathbf{z}_i, \mathbf{z}_{<i})\approx \mathbf{x}_i$. This reconstruction path constrains $\mathbf{z}_i$ to retain the information needed for pixel-space output, while allowing the AR model to generate in a lower-dimensional space.

A straightforward way to obtain such low-dimensional targets is to use an external VAE or tokenizer. However, this would introduce a separate representation-learning stage and turn the model into a two-stage latent AR system. We instead learn the intermediate targets end-to-end with the AR model. The simplest end-to-end construction is to map the current ground-truth token alone, e.g., $\mathbf{z}_i=g_\psi(\mathbf{x}_i)$. However, such a target is purely local and ignores the causal prefix representation already computed by the AR Transformer.

PRA therefore constructs context-aware intermediate targets. Under teacher forcing, the AR Transformer encodes the ground-truth prefix and produces the causal representation $\mathbf{h}_{i-1}=f_\theta(\mathbf{x}_{<i})$. We combine this prefix representation with the current ground-truth token through a lightweight projection network:
\begin{equation}
    \mathbf{z}_i = g_\psi(\mathbf{x}_i, \mathbf{h}_{i-1}), 
    \quad \mathbf{z}_i \in \mathbb{R}^{d_z}, \quad d_z < d .
    \label{eq:latent}
\end{equation}
Here, $\mathbf{x}_i$ is used only to define the training target $\mathbf{z}_i$; the AR model itself must generate this state from the causal prefix. Using $\mathbf{h}_{i-1}$ introduces little additional cost, since the prefix representation is already available from the teacher-forced AR forward. To prevent $g_\psi$ from relying only on the current token, we apply token masking: with probability $p_{\mathrm{mask}}$, $\mathbf{x}_i$ is replaced by a learnable mask embedding before projection, encouraging $\mathbf{z}_i$ to incorporate causal context.

Given these intermediate targets, the token-level diffusion head is trained in the low-dimensional space using $\mathbf{z}_i$ as the target state. During generation, the AR model generates an intermediate state $\hat{\mathbf{z}}_i$, and the pixel decoder maps it to the pixel-space token $\hat{\mathbf{x}}_i=D_\phi(\hat{\mathbf{z}}_i, \hat{\mathbf{z}}_{<i})$. Thus, PRA reduces the difficulty of each AR generation step while preserving a pixel-in, pixel-out interface.

\subsection{Parallel Construction of Inference-Like Pixel Inputs}
\label{sec:pra_inputs}

Low-dimensional intermediate targets reduce the output-side difficulty of generating each token, but they do not by themselves resolve the input-side train--inference gap. If the AR model is still trained only on clean ground-truth pixel prefixes, it may remain brittle when conditioned on decoded tokens produced by its own generations during inference-time rollout. PRA therefore approximates this rollout pixel-input interface in parallel. Rather than sampling a full autoregressive trajectory, it constructs decoded pixel inputs independently at each position through the same intermediate-state-to-pixel path used at inference.

For each target $\mathbf{z}_i$, we sample $\mathbf{z}_i^0\sim\mathcal{N}(\mathbf{0},\mathbf{I})$ and $t_i\sim\mathcal{U}(t_{\min},1)$, and form a perturbed state
\begin{equation}
    \tilde{\mathbf{z}}_i=(1-t_i)\mathbf{z}_i^0+t_i\mathbf{z}_i .
\end{equation}
The causal pixel decoder then maps this perturbed state back to pixel space:
\begin{equation}
    \bar{\mathbf{x}}_i = D_\phi(\tilde{\mathbf{z}}_i, \tilde{\mathbf{z}}_{<i}).
\end{equation}
The resulting token $\bar{\mathbf{x}}_i$ is not sampled from an actual AR trajectory. Instead, it serves as an inference-like pixel input because it is produced through the same intermediate-state-to-pixel decoding path used at inference time. This makes it closer to inference-time decoded tokens than either clean ground-truth inputs or independent pixel-space noise injection, and we provide supporting evidence for this design through input-side ablations in Sec.~\ref{sec:input_ablation} and alignment diagnostics in Appendix~\ref{app:rollout_diagnostics}.

The reconstructed sequence $\bar{\mathbf{x}}=(\bar{\mathbf{x}}_1,\ldots,\bar{\mathbf{x}}_T)$ is then used as a stop-gradient input sequence for AR training. With the standard causal shift, the hidden state for position $i$ is computed from the reconstructed prefix $\bar{\mathbf{x}}_{<i}$ rather than the clean prefix $\mathbf{x}_{<i}$, and the token-level diffusion head is trained to generate the target intermediate state $\mathbf{z}_i$. Since perturbed intermediate states at all positions are available during training, the pixel decoder can construct the decoded inputs in parallel under causal masking. In practice, PRA requires an additional parallel AR forward during training, but avoids the sequential sampling cost. \looseness=-1

\subsection{Overall Training and Inference Pipeline}
\label{sec:pra_pipeline}

\begin{figure}[t]
    \centering

    \begin{minipage}[t]{0.51\textwidth}
    \vspace{0pt}
    \begin{algorithm}[H]
    \caption{PRA training}
    \label{alg:train}
    \centering
    \begin{algorithmic}[1]
    \small
    \Require image $\mathtt{IMG}$, pixel tokens $\mathbf{x}_{1:T}$, condition $c$
    \Require AR Transformer $f_\theta$, target encoder $g_\psi$, causal pixel decoder $D_\phi$,
    flow head $v_\omega$
    \Statex

    \For{$i=1,\dots,T$ \textbf{in parallel}}
      \State $\mathbf{h}_{i-1} \leftarrow f_\theta(\mathbf{x}_{<i})$
      \State $\mathbf{z}_i\gets g_\psi(\mathbf{x}_i,\mathbf{h}_{i-1})$
      \Comment{low-dim inter. target}
      \State sample $\mathbf{z}_i^0\sim\mathcal{N}(\mathbf{0},\mathbf{I})$,
      $t_i\sim\mathcal{U}(t_{\min},1)$
      \State $\tilde{\mathbf{z}}_i\gets(1-t_i)\mathbf{z}_i^0+t_i\mathbf{z}_i$
      \State $\bar{\mathbf{x}}_i\gets D_\phi(\tilde{\mathbf{z}}_i, \tilde{\mathbf{z}}_{<i})$
      \Comment{decoded rollout-like pixel token}
    \EndFor
    \Statex

    \For{$i=1,\dots,T$ \textbf{in parallel}}
      \State $\bar{\mathbf{h}}_{i-1} \leftarrow f_\theta(c, \operatorname{sg}(\bar{\mathbf{x}}
      _{<i}))$
      \State sample $\epsilon_i \sim \mathcal{N}(\mathbf{0},\mathbf{I})$,
      $s_i \sim \operatorname{LogitNormal}(0,1)$
      \State $\mathbf{z}_i^{s_i} \leftarrow (1-s_i)\epsilon_i + s_i \mathbf{z}_i$
      \Comment{Flow Matching train}
    \EndFor
    \State $\mathcal{L}_{\rm AR}\gets\frac{1}{T}\sum_{i=1}^T\left\|v_\omega(\mathbf{z}_i^{s_i}, s_i, \bar{\mathbf{h}}_{i-1})-(\mathbf{z}_i-\epsilon_i)\right\|_2^2$

    \State $\mathcal{L}_{\rm rec}\gets
    \ell_{\rm rec}\!\left(
    \mathrm{Unpatchify}(\bar{\mathbf{x}}_{1:T}),\mathtt{IMG}
    \right)$
    \State $\mathcal{L}\gets
    \mathcal{L}_{\rm AR}+\mathcal{L}_{\rm rec}+\mathcal{L}_{\rm aux}$
    \State \Return $\mathcal{L}$
    \end{algorithmic}
    \end{algorithm}
    \end{minipage}
    \hfill
    \begin{minipage}[t]{0.46\textwidth}
    \vspace{0pt}
    \begin{algorithm}[H]
    \caption{PRA inference: sequential rollout}
    \label{alg:inference}
    \centering
    \begin{algorithmic}[1]
    \small
    \Require condition $c$
    \Require $f_\theta$, $D_\phi$, $v_\omega$

    \vspace{0.8\baselineskip} 
    \State $\hat{\mathbf{x}}_{<1}\gets\emptyset$

    \For{$i=1,\dots,T$}
      \State $\hat{\mathbf{h}}_{i-1}\gets f_\theta(c,\hat{\mathbf{x}}_{<i})$
      \Comment{generated prefix}
      \vspace{1.7\baselineskip} 
      \State $\hat{\mathbf{z}}_i\gets
      \textsc{FlowSample}(v_\omega,\hat{\mathbf{h}}_{i-1})$
      \State $\hat{\mathbf{x}}_i\gets D_\phi(\hat{\mathbf{z}}_i, \hat{\mathbf{z}}_{<i})$
      \Comment{decoded rollout pixel token}
      \State append $\hat{\mathbf{x}}_i$ to the prefix
      \Statex
    \EndFor
    \Statex

    \State \Return $\mathrm{Unpatchify}(\hat{\mathbf{x}}_{1:T})$
    \end{algorithmic}
    \end{algorithm}
    \end{minipage}
  \end{figure}

Alg.~\ref{alg:train} and ~\ref{alg:inference} summarize the overall training and inference pipeline of PRA. During training, PRA first runs a teacher-forced AR forward on the ground-truth pixel sequence to obtain the causal representations used for constructing the target intermediate states $\mathbf{z}_i$. These target states are then perturbed and passed to the pixel decoder, which reconstructs the corresponding pixel tokens and produces inference-like pixel inputs $\bar{\mathbf{x}}_i$. In a second parallel AR forward, the model conditions on the reconstructed prefix $\bar{\mathbf{x}}_{<i}$ and trains the token-level diffusion head to generate the target state $\mathbf{z}_i$. Thus, the AR model learns to generate intermediate states under imperfect pixel prefixes, while the pixel decoder learns to map perturbed intermediate states back to pixel space.

During inference, PRA uses the same intermediate-state-to-pixel path autoregressively. At each step, the model conditions on previously generated pixel tokens $\hat{\mathbf{x}}_{<i}$, generates a low-dimensional intermediate state $\hat{\mathbf{z}}_i$, decodes it into the next pixel token $\hat{\mathbf{x}}_i=D_\phi(\hat{\mathbf{z}}_i, \hat{\mathbf{z}}_{<i})$, and feeds this token back for subsequent generation. Therefore, PRA keeps the external AR interface pixel-in and pixel-out, while using low-dimensional intermediate states only internally.

\vspace{-4pt}
\subsection{Discussion}
\vspace{-6pt}

\paragraph{Pixel-space AR vs. diffusion models.}
Pixel-space diffusion models such as JiT also model high-dimensional pixel patches, but they avoid the autoregressive loop that makes local errors accumulate across spatial positions. In pixel-space AR, each generated patch is fed back into the causal context for subsequent predictions, so single-step errors can become non-local and progressively degrade later patches. This makes train--inference mismatch a key bottleneck for pixel-space AR, even when using $x$-prediction that works well for diffusion models. PRA reuses the pixel decoder to construct rollout-like pixel inputs in parallel, while also reducing single-step difficulty through low-dimensional intermediate states, thereby narrowing the gap to diffusion-based approaches.

\vspace{-8pt}
\paragraph{PRA vs. two-stage latent AR models.}
Two-stage latent AR models reduce generation difficulty by relying on an external tokenizer or autoencoder, often with AR-specific regularization of the latent space. PRA also introduces low-dimensional intermediate states, but learns them end-to-end with the AR model and uses them only as internal generation targets. This removes the need for a separately pretrained tokenizer while preserving a pixel-in, pixel-out AR interface: the model takes pixel patches as inputs and generates pixel patches as outputs. In pixel-space modeling, this unified raw-token interface can benefit both image generation and visual understanding, as reflected by PRA's stronger image classification probing performance compared with latent-space AR baselines (Sec.~\ref{sec:linearprob}).

\vspace{-4pt}
\section{Experiments}
\vspace{-6pt}

We conduct a systematic evaluation of PRA on class-conditional ImageNet-1K generation at $256\times256$ resolution, using raw pixel patches as continuous tokens. We first compare PRA against strong baselines from two-stage latent-space generation, pixel-space diffusion, and pixel-space AR. We then validate the two main components of PRA, end-to-end intermediate targets for pixel outputs and parallel inference-like pixel inputs, through output-side and input-side ablations. Finally, beyond generation quality, we evaluate image understanding through ImageNet classification probing.

\vspace{-4pt}
\subsection{Implementation Details}
\vspace{-6pt}

We evaluate three model scales: PRA-S (135M), PRA-B (250M), and PRA-L (511M). All variants use the same PRA architecture and differ only in model depth and width, with architecture details provided in Appendix~\ref{app:details}. We use $16\times16$ pixel patches, each flattened into a $d=16\times16\times3=768$-dimensional continuous token, yielding 256 tokens per $256\times256$ image. Following prior work~\citep{ke2026hyperspherical}, we use 16 class-conditioning prefix tokens with dropout probability $0.1$. Unless otherwise specified, we set the intermediate-state dimension to $d_z=16$, the noise lower bound to $t_{\min}=0.5$, and the token masking probability to $p_{\mathrm{mask}}=0.5$.

All main models are trained end-to-end on ImageNet-1K for 400 epochs. Unless otherwise specified, ablation experiments use the PRA-B scale and are trained for 100 epochs. The training objective combines a rectified-flow loss for AR generation in the intermediate-state space and a reconstruction loss for the pixel decoder. The reconstruction loss consists of an $\ell_1$ pixel loss and an LPIPS loss on reconstructed images~\citep{zhang2018unreasonable}. We use equal weights for all loss terms without tuning. Optimization uses AdamW~\citep{kingma2014adam,loshchilov2017decoupled} with batch size 512, $\beta=(0.9,0.95)$, weight decay 0.05, a cosine learning-rate schedule with 20K warmup steps, peak learning rate $3\times10^{-4}$, and exponential moving average decay 0.9999. During inference, we generate each intermediate state with the diffusion head using 100 Euler--Maruyama steps followed by a final deterministic Euler step, as detailed in Appendix~\ref{app:sampling_details}. We use the linear classifier-free guidance (CFG) schedule from MAR and enable KV caching for efficient autoregressive generation. \looseness=-1

\begin{table}[t]
\centering
\small
\setlength{\tabcolsep}{3pt}
\renewcommand{\arraystretch}{1.05}

\begin{minipage}[t]{0.55\linewidth}
\vspace{0pt}
\centering

\captionof{table}{Comparison of image generation performance on class-conditional ImageNet-1K at 256$\times$256 resolution, reporting FID and IS.}
\vspace{3pt}
\label{tab:main}

\small
\setlength{\tabcolsep}{0.7pt}
\renewcommand{\arraystretch}{1.0}

\begin{tabular}{@{}lccc@{}}
\toprule
\textbf{Model} &
\textbf{Params} &
\textbf{FID}$\downarrow$ &
\textbf{IS}$\uparrow$ \\
\midrule

\rowcolor[gray]{0.9}\multicolumn{4}{@{}l}{\textit{Two-Stage Diffusion}} \\
DiT-XL/2 \scitep{Peebles_2023_ICCV} & 675+49M & 2.27 & 278.2 \\
SiT-XL/2 \scitep{ma2024sit} & 675+49M & 2.06 & 277.5 \\
REPA-XL/2 \scitep{yu2024representation} & 675+49M & 1.42 & 305.7 \\
LightningDiT-XL/2 \scitep{yao2025reconstruction} & 675+49M & 1.35 & 295.3 \\
DDT-XL/2 \scitep{wang2025ddt} & 675+49M & 1.26 & 310.6 \\
RAE-XL/2 \scitep{zheng2025diffusion} & 839+415M & \textbf{1.13} & 262.6 \\
\midrule

\rowcolor[gray]{0.9}\multicolumn{4}{@{}l}{\textit{Two-Stage Autoregressive}} \\
VAR-d20 \scitep{tian2024visual} & 600M+40M & 2.57 & 302.6 \\
LlamaGen-XL \scitep{sun2024autoregressive} & 1.4B+70M & 2.34 & 253.9 \\
MAR-B \scitep{li2024autoregressive} & 208+49M & 2.31 & 281.7 \\
MAR-L \scitep{li2024autoregressive} & 479+49M & 1.78 & 296.0 \\
SphereAR-L \scitep{ke2026hyperspherical} & 479+49M & \textbf{1.54} & 295.9 \\
\midrule

\rowcolor[gray]{0.9}\multicolumn{4}{@{}l}{\textit{Pixel-Space Diffusion}} \\
ADM-G \scitep{dhariwal2021diffusion} & 554M & 4.59 & 186.7 \\
SiD \scitep{hoogeboom2023simple} & 2B & 2.44 & 256.3 \\
SiD2 \scitep{hoogeboom2025simpler} & N/A & \textbf{1.38} & -- \\
PixelFlow-XL/4 \scitep{chen2025pixelflow} & 677M & 1.98 & 282.1 \\
PixNerd-L/16 \scitep{wang2025pixnerd} & 458M & 2.64 & 297.0 \\
JiT-B/16 \scitep{li2025back} & 131M & 3.66 & 275.1 \\
JiT-L/16 \scitep{li2025back} & 459M & 2.36 & 298.5 \\
\midrule

\rowcolor[gray]{0.9}\multicolumn{4}{@{}l}{\textit{Pixel-Space Autoregressive}} \\
JetFormer \scitep{tschannen2025jetformerautoregressivegenerativemodel} & 2.8B & 6.64 & -- \\
FractalMAR-B \scitep{li2025fractal} & 186M & 11.8 & 274.3 \\
FractalMAR-L \scitep{li2025fractal} & 438M & 7.30 & 334.9 \\
FractalMAR-H \scitep{li2025fractal} & 848M & 6.15 & 348.9 \\
FARMER-1.1B/8 \scitep{zheng2025farmer} & 1.1B & 5.02 & 237.0 \\
FARMER-1.9B/8 \scitep{zheng2025farmer} & 1.9B & 3.60 & 269.2 \\
\textbf{PRA-S (ours)} & 135M & 2.58 & 273.9 \\
\textbf{PRA-B (ours)} & 250M & 2.21 & 276.9 \\
\textbf{PRA-L (ours)} & 511M & \textbf{1.94} & 287.3 \\
\bottomrule
\end{tabular}
\end{minipage}
\hfill
\begin{minipage}[t]{0.42\linewidth}
  \centering
  \setlength{\tabcolsep}{3pt}
  \renewcommand{\arraystretch}{1.0}

    \captionof{table}{Ablations on learned intermediate targets.}
  \label{tab:intermediate_ablation}

  \begin{tabular}{llcc}
  \toprule
  \textbf{Factor} & \textbf{Variant} & \textbf{FID}$\downarrow$ & \textbf{IS}$\uparrow$ \\
  \midrule
  \multirow{3}{*}{Target}
    & LDM enc. & 3.37 & 266.73 \\
    & Local only & 3.08 & 271.96 \\
    & \textbf{Prefix aware} & \textbf{2.88} & 279.05 \\
    
  \midrule
  \multirow{4}{*}{Masking}
    & $p_{\mathrm{mask}}=0.0$ & 2.96 & 280.32 \\
    & $p_{\mathrm{mask}}=0.3$ & \textbf{2.79} & 280.66 \\
    & $p_{\mathrm{mask}}=0.5$ & 2.88 & 279.05 \\
    & $p_{\mathrm{mask}}=0.7$ & 3.01 & 279.03 \\
  
  \midrule
  \multirow{4}{*}{Dim.}
    & $d_z=8$ & 3.36 & 285.85 \\
    & \textbf{$d_z=16$} & \textbf{2.88} & 279.05 \\
    & $d_z=32$ & 3.50 & 297.17 \\
    & $d_z=64$ & 7.03 & 220.11 \\
  \bottomrule
  \end{tabular}%

  \vspace{8pt}
  \captionof{table}{Ablations on AR inputs with low-dimensional prediction targets.}
  \label{tab:input_ablation}
    \vspace{-5pt}
  \begin{tabular}{lcc}
  \toprule
  Input & FID $\downarrow$ & IS $\uparrow$ \\
  \midrule
  GT pixel & 42.36 & 72.50 \\
  GT pixel + noise & 32.60 & 110.60 \\
  GT Inter. states & 3.21 & 261.86 \\
  GT Inter. states + noise & 3.05 & 268.34 \\
  \textbf{Decoded Pixel (PRA)} & \textbf{2.88} & 279.05 \\
  \bottomrule
  \end{tabular}%

  \vspace{8pt}
  
  \captionof{table}{Ablations on robust pixel inputs.}
  \label{tab:t_min}
    \vspace{-5pt}
  \begin{tabular}{lcc}
  \toprule
  Noise level & FID $\downarrow$ & IS $\uparrow$ \\
  \midrule
  $t_{\min}=1.0$ & 3.35 & 283.60 \\
  \textbf{$t_{\min}=0.7$} & \textbf{2.62} & 282.03 \\
  $t_{\min}=0.5$ & 2.88 & 279.05 \\
  $t_{\min}=0.3$ & 2.96 & 291.34 \\
  $t_{\min}=0.0$ & 3.06 & 291.01 \\
  \bottomrule
  \end{tabular}%

\end{minipage}
\par\vspace{-10pt} 
\end{table}

\begin{table}[t]
  \centering
  \small
  \setlength{\tabcolsep}{3pt}
  \begin{minipage}[t]{0.46\linewidth}
    \centering
    \vspace{0pt}
    \caption{ImageNet linear probing results. PRA-L achieves highest top-1 accuracy.}
    \label{tab:probing}
  \end{minipage}\hfill
  \begin{minipage}[t]{0.52\linewidth}
    \centering
    \vspace{0pt}
    \caption{Inference efficiency. Throughput parentheses show speedup over the corresponding JiT baseline; GFLOPs include CFG and show percentages.}
    \label{tab:inference_efficiency}
  \end{minipage}

  \begin{minipage}[t]{0.46\linewidth}
    \centering
    \vspace{0pt}
    \resizebox{\linewidth}{!}{%
    \begin{tabular}{lllcc}
    \toprule
    Paradigm & Space & Model & Param. & \shortstack{Top-1 Acc.(\%) $\uparrow$} \\
    \midrule
    \multirow{2}{*}{Diffusion}
      & Latent & DiT-XL/2 & 675+49M & 43.28 \\
      & Pixel  & JiT-L & 459M & 42.76 \\
    \midrule
    \multirow{2}{*}{AR}
      & Latent & SphereAR-L & 479+49M & 52.19 \\
      & Pixel  & \textbf{PRA-L} & 511M & \textbf{68.80} \\
    \bottomrule
    \end{tabular}%
    }
  \end{minipage}\hfill
  \begin{minipage}[t]{0.52\linewidth}
    \centering
    \vspace{0pt}
    \resizebox{\linewidth}{!}{%
    \begin{tabular}{lrlrrr}
        \toprule
        Model & Params & Sampler & Batch &
        \shortstack{GFLOPs/image$\downarrow$} &
        \shortstack{Throughput\\(images/s)$\uparrow$} \\
        \midrule
        JiT-B/16 & 131M & Heun 50 & 896 &
        \shortstack{9,988(100\%)} & \shortstack{6.99 ($1.00\times$)} \\
        PRA-S & 135M & EM 100 & 896 &
        \shortstack{\textbf{1,655}\textbf{(16.6\%)}} &
        \shortstack{\textbf{33.46} ($\mathbf{4.79\times}$)} \\
        \midrule
        JiT-L/16 & 459M & Heun 50 & 292 &
        \shortstack{34,826(100\%)} & \shortstack{2.13 ($1.00\times$)} \\
        PRA-L & 511M & EM 100 & 292 &
        \shortstack{\textbf{5,478}\textbf{(15.7\%)}} &
        \shortstack{\textbf{7.78} ($\mathbf{3.66\times}$)} \\
        \bottomrule
    \end{tabular}%
    }
  \end{minipage}
  \par\vspace{-10pt} 
\end{table}

\vspace{-4pt}
\subsection{Overall Image Generation Performance}
\vspace{-6pt}

Following prior work~\citep{li2025back}, we generate 50K samples and report Fr\'{e}chet Inception Distance (FID)~\citep{heusel2017gans} as the primary fidelity metric, together with Inception Score (IS)~\citep{salimans2016improved}. All metrics are computed using the ADM evaluation suite~\citep{dhariwal2021diffusion} against the standard ImageNet-256 reference statistics. For each model, we tune the classifier-free guidance (CFG) scale with a step size of 0.1 and report the best result.

We compare PRA with four families of methods along two axes: pixel-space versus two-stage generation, depending on whether a pretrained tokenizer or autoencoder is required, and autoregressive versus diffusion generation, depending on the generation paradigm. PRA falls into the pixel-space autoregressive family: it retains pixel-space feedback and learns its intermediate representations jointly, without a separately pretrained tokenizer.

Table~\ref{tab:main} summarizes the results. PRA achieves a substantially better trade-off between generation quality and model scale than prior pixel-space AR methods. PRA-S, with only 135M parameters, reaches an FID of 2.58, outperforming much larger pixel-space AR models such as FARMER-1.9B/8, which has 1.9B parameters and an FID of 3.60. PRA also improves consistently with scale: PRA-B reaches an FID of 2.21, and PRA-L further improves it to 1.94. This establishes a new state of the art among pixel-space AR models, reducing the best reported FID in this family from 3.60 to 1.94. Compared with two-stage AR methods and pixel-space diffusion models, PRA remains competitive while directly modeling raw pixel tokens. These results show that PRA substantially strengthens pixel-space AR, making it competitive with strong two-stage and diffusion-based image generators.

\vspace{-4pt}
\subsection{Output-Side Ablations: Learning AR-aligned Intermediate Targets}
\label{sec:output_ablation}
\vspace{-6pt}

We first study the output-side component of PRA: replacing direct prediction of high-dimensional pixel patches with learned intermediate states. The goal is not to learn a generic image latent space, but to construct an intermediate target space that is easier for the current causal AR model to predict while preserving enough information to decode back to pixel patches. Table~\ref{tab:intermediate_ablation} ablates how the target space is constructed and how large the intermediate states should be, with all variants using decoded pixel inputs for AR training.

For target construction (Table~\ref{tab:intermediate_ablation}), a frozen LDM encoder~\citep{rombach2022high} performs worse than the learned PRA target, indicating that an off-the-shelf latent space is not necessarily suitable for causal AR prediction. A local-only target, encoding only the current pixel patch, also underperforms the prefix-aware target. The prefix-aware target incorporates the causal prefix representation $\mathbf{z}_i=g_\psi(\mathbf{x}_i,\mathbf{h}_{i-1})$ and performs best, improving FID from 3.08 to 2.88, showing that the intermediate target should be adapted to the causal context, rather than serving merely as a local autoencoding code. Encoder masking further regularizes this target construction. Without masking, the encoder can rely too heavily on the clean current patch, weakening the target's dependence on the causal prefix. Moderate token masking encourages the intermediate state to use prefix information and improves its compatibility with AR prediction, while overly strong masking removes too much local information and makes decoding harder. We therefore use the prefix-aware target with $p_{\rm mask}=0.5$ by default.

We next vary the intermediate-state dimension $d_z$, which controls the trade-off between predictability and information preservation. If the state is too small, such as $d_z=8$, it lacks sufficient capacity to reconstruct pixel patches. If the state is too large, the AR model again faces a harder continuous-token prediction problem. This is reflected by the degradation at $d_z=32$ and especially $d_z=64$. The best performance is obtained at $d_z=16$, suggesting that a compact but expressive intermediate space is crucial for reducing single-step errors while maintaining pixel decodability.

\vspace{-4pt}
\subsection{Input-Side Ablations: Inference-Like Pixel Inputs}
\label{sec:input_ablation}
\vspace{-6pt}

We next ablate the AR training inputs, which target the input-side train--inference mismatch. Taking low-dimensional token compression as a starting point, we retain low-dimensional prediction targets and vary the AR inputs in Table~\ref{tab:input_ablation}. Using clean ground-truth (GT) pixels as inputs performs poorly (FID 42.36), even though the output target is already low-dimensional. This is because inference does not condition on ground-truth pixels, but on pixel tokens decoded from generated intermediate states. These decoded pixels can deviate substantially from real pixels, creating a strong input mismatch. Pixel-space noise injection improves FID to 32.60, but independent perturbations still do not match the structured artifacts introduced by the intermediate-state-to-pixel path.

We then consider using ground-truth(GT) intermediate states as AR inputs. Without noise, this variant still suffers from a mismatch, since training conditions on target states constructed from GT tokens, whereas inference must rely on states produced by the model. Adding noise to the GT intermediate states makes the inputs closer to generated states and further improves performance, indicating that the intermediate space is easier for autoregressive modeling. However, this variant changes the external AR interface from pixels to learned-states, making the model closer to latent-space AR.

In contrast, Decoded Pixel (PRA) preserves the pixel-in, pixel-out interface while better matching the inference condition. It constructs training inputs by perturbing intermediate states and mapping them through the same pixel decoder used at inference. This provides a parallel approximation to autoregressive rollout inputs: each position is constructed independently, but through the same intermediate-state-to-pixel path used during sequential inference. As a result, the AR model is trained on decoded pixel inputs that more closely resemble the feedback it receives during inference. PRA achieves the lowest FID in Table~\ref{tab:input_ablation} (2.88), outperforming variants that use clean pixels, independently noised pixels, or GT intermediate states as AR inputs.

We further study the perturbation strength used to construct decoded pixel inputs in Table~\ref{tab:t_min}. The lower bound $t_{\min}$ controls how close the perturbed state is to the target intermediate state: larger values produce cleaner inputs, while smaller values introduce stronger perturbations. Both extremes can be suboptimal. If the inputs are too clean, training remains close to teacher forcing; if they are too corrupted, the causal context becomes unreliable. A moderate value provides the best trade-off between inference-likeness and prefix reliability. We use $t_{\min}=0.5$ unless otherwise specified.


\vspace{-4pt}
\subsection{Image Understanding via Linear Probing}
\label{sec:linearprob}
\vspace{-6pt}

A potential advantage of pixel-space autoregressive modeling is that the model operates directly on raw visual tokens, which may preserve useful information for visual understanding. We evaluate this by applying the MAE-style ImageNet-1K linear probing protocol~\citep{he2022mae} to the generation-trained backbone. The pretrained model is frozen, and only a linear classifier is trained.

We evaluate PRA-L directly, without architectural modifications or additional unsupervised pretraining. As shown in Table~\ref{tab:probing}, PRA-L achieves a top-1 accuracy of 68.80\%, substantially outperforming both the latent-space AR baseline SphereAR-L and the pixel-space diffusion JiT-L. This suggests that PRA not only improves pixel-space generation, but also learns more transferable visual representations, supporting its potential as a unified pixel-space model for image generation and understanding.

\vspace{-4pt}
\subsection{Inference Efficiency}
\label{sec:efficiency}
\vspace{-6pt}

With KV caching, PRA achieves substantially higher generation throughput than the pixel-space diffusion baseline JiT. We evaluate both methods at comparable parameter scales using their default samplers on a single RTX 4090 (24 GB) at $256\times256$ resolution, with matched batch sizes within each pair. As shown in Table~\ref{tab:inference_efficiency}, PRA-S and PRA-L achieve $4.79\times$ and $3.66\times$ the generation throughput of their respective JiT counterparts and require only $16.6\%$ and $15.7\%$ of the FLOPs.




\vspace{-8pt}
\section{Conclusion}
\vspace{-6pt}

We presented \emph{Parallel Rollout Approximation} (PRA) for pixel-space AR image generation. PRA addresses two coupled challenges of high-dimensional continuous-token AR: large single-step errors from directly generating pixel patches, and cross-step error accumulation caused by train--inference mismatch. PRA learns low-dimensional intermediate targets end-to-end to reduce output-side generation difficulty. The pixel decoder both maps generated states to inference-time pixel outputs and constructs training inputs from perturbed target states, mitigating the input-side mismatch while preserving parallel training and a pixel-in, pixel-out AR interface. Experiments on class-conditional ImageNet-1K generation show that PRA substantially improves pixel-space AR generation and establishes a new SOTA among pixel-space AR models. Beyond generation, linear probing results suggest that PRA also learns transferable visual representations, supporting its potential for unified pixel-space image generation and understanding.

\textbf{Limitations.}
PRA introduces several additional components. Although they are trained end-to-end without a separately pretrained tokenizer, simplifying the framework and validating it across broader data domains and generation tasks remain important directions for future work.

\clearpage
\begin{ack}
DH is supported by National Science and Technology Major Project (2022ZD0114902), National Science Foundation of China (NSFC62376007), National Science Foundation of China (under Key Project No. 92570203), Beijing Natural Science Foundation (Z250001), and Beijing Major Science and Technology Project under Contract no. Z251100008425004. This work is supported in part by the State Key Laboratory of General Artificial Intelligence.
\end{ack}

\bibliographystyle{plainnat}
\bibliography{references}


\newpage
\appendix

\section{Model Details} \label{app:details}
\vspace{-15pt}

\begin{table}[h]
\centering
\small
\setlength{\tabcolsep}{4pt}
\caption{PRA scale configurations. ``res / AdaLN'' denotes the number of
diffusion-head residual blocks and the number of shared AdaLN groups. The
diffusion head width matches $d$. The pixel-decoder width $d_r$ also defines the
encoder $g_\psi$ width. All blocks use an MLP ratio of 4 with SwiGLU.
Training cost is reported as wall-clock days on 8$\times$A100 GPUs for 400-epoch training.}
\label{tab:scales}
\begin{tabular}{lcccccccccc}
\toprule
Model & \multicolumn{3}{c}{AR Transformer $f_\theta$}
      & \multicolumn{2}{c}{Diffusion head $v_\omega$}
      & \multicolumn{3}{c}{Pixel decoder $D_\phi$}
      & Params. & Training Cost by \\
\cmidrule(lr){2-4}\cmidrule(lr){5-6}\cmidrule(lr){7-9}
 & layers & heads & $d$
 & res / AdaLN & width
 & layers & heads & $d_r$
 & & 8$\times$A100 (days) \\
\midrule
PRA-S  & 12 & 12 & 768  & 4 / 1 & 768  & 6 & 8  & 512 & 135M & 3.125 \\
PRA-B  & 24 & 12 & 768  & 6 / 2 & 768  & 6 & 12 & 768 & 250M & 6 \\
PRA-L  & 30 & 16 & 1024 & 8 / 2 & 1024 & 8 & 12 & 768 & 511M & 14.3 \\
\bottomrule
\end{tabular}
\end{table}

\subsection{Architecture Overview}

PRA consists of four modules, all trained end-to-end: a causal AR Transformer $f_\theta$, an intermediate-state encoder $g_\psi$ that produces $\mathbf{z}_i \in \mathbb{R}^{d_z}$, a pixel decoder $D_\phi$ that maps a noisy intermediate state back to a continuous pixel patch, and a diffusion head $v_\omega$ that models the rectified-flow distribution of $\mathbf{z}_i$ conditioned on the AR hidden state. The input image is patchified into non-overlapping $16{\times}16$ patches in raster-scan order. For a $256{\times}256$ RGB image, this yields $T=256$ tokens, each corresponding to a $16{\times}16{\times}3=768$-dimensional pixel patch. The scale-specific architectural configurations and training costs are summarized in Table~\ref{tab:scales}.

\subsection{Causal AR Transformer \texorpdfstring{$f_\theta$}{f}}

The backbone is a causal Transformer~\citep{vaswani2017attention} with pre-RMSNorm attention~\citep{zhang2019root}, SwiGLU feed-forward layers~\citep{shazeer2020glu}, and 2-D rotary position embeddings (RoPE)~\citep{su2024roformer} over patch coordinates. Class conditioning is injected through $K{=}16$ learnable prefix tokens, obtained from an embedding table indexed by the class label and prepended to the patch sequence. We use FlashAttention~\citep{dao2022flashattention} during training. Class dropout with probability $0.1$ replaces the class label with a null label, enabling classifier-free guidance at inference.

\subsection{Intermediate-State Encoder \texorpdfstring{$g_\psi$}{g}}

The intermediate-state encoder maps each pair $(\mathbf{x}_i, \mathbf{h}_{i-1})$ to a low-dimensional intermediate state $\mathbf{z}_i$. We use $d_z{=}16$ for all scales. The encoder first linearly projects the pixel patch to the encoder width $d_r$, then applies four SwiGLU residual blocks with LayerNorm and AdaLN modulation conditioned on $\mathbf{h}_{i-1}$. The conditioning representation is the AR hidden state after RMSNorm and an additive learnable position embedding. AdaLN parameters are shared every two residual blocks. The encoder hidden width matches the pixel-decoder width $d_r$ in Table~\ref{tab:scales}.

To encourage $\mathbf{z}_i$ to depend on causal context rather than only the current patch, we use two-level encoder masking during training. With probability $p_{\rm sample}$, a training example is selected for masking; within selected examples, each token is replaced by a learned mask embedding with probability $p_{\rm token}$. Unless otherwise specified, we use $p_{\rm sample}{=}0.9$ and $p_{\rm token}{=}0.5$.

\subsection{Pixel Decoder \texorpdfstring{$D_\phi$}{D}}

The pixel decoder is a causal Transformer that maps noisy intermediate states back to pixel patches. It is one-step in the sense that reconstruction uses a single forward pass without iterative refinement; the diffusion head separately uses multiple sampling steps to generate each intermediate state. It consists of three components: (i) a linear projection from $\mathbb{R}^{d_z}$ to $\mathbb{R}^{d_r}$, (ii) a causal Transformer with the same block design as the AR backbone but separate weights, with depth and width specified in Table~\ref{tab:scales}, and (iii) an RMSNorm followed by a two-layer SiLU MLP head that projects the hidden representation back to a $16{\times}16{\times}3=768$-dimensional pixel patch. Causality is preserved so that KV caching can be used; at inference, $D_\phi$ is invoked once per autoregressive step. The final output linear layer is zero-initialized.

\subsection{Diffusion Head \texorpdfstring{$v_\omega$}{v}}

The diffusion head models the rectified-flow distribution of $\mathbf{z}_i$ conditioned on the AR hidden state. It is implemented as a SwiGLU MLP stack with shared AdaLN modulation. The conditioning vector is $\mathbf{c}_i=\operatorname{RMSNorm}(\mathbf{h}_{i-1})+\mathbf{p}^{\rm diff}_i$, where $\mathbf{p}^{\rm diff}_i$ is a learned position embedding. For flow timestep $s$, the modulation input is $\mathrm{SiLU}(t_{\text{embed}}(s) + \mathbf{c}_i)$, where $t_{\text{embed}}$ is a sinusoidal timestep embedding. The head predicts the rectified-flow velocity. Both the final output linear layer and the AdaLN modulation linear layer are zero-initialized.

\section{Training Objective and Implementation Details}
\label{app:pra_training}

PRA uses parallel training while exposing the AR model to inputs that better match inference-time generated tokens. The implementation performs a reconstruction pass and an AR training pass. In the reconstruction pass, ground-truth pixel patches are encoded into intermediate states, optionally perturbed in intermediate space, decoded back to pixel patches, and supervised by image reconstruction losses. In the AR pass, the decoded pixel patches are stop-gradient inputs to the causal AR Transformer, and the diffusion head is trained to generate the clean intermediate targets.

\paragraph{Intermediate-space perturbation.}
During training, the decoder input is a perturbed version of the target intermediate state. Perturbation times are sampled independently for each token. For selected samples, we draw $\epsilon_i\sim\mathcal{N}(0,I)$ and $t_i\sim\mathcal{U}(t_{\min},1)$ and form
\begin{equation}
    \tilde{\mathbf{z}}_i = t_i \mathbf{z}_i + (1-t_i)\epsilon_i .
\end{equation}
Unless otherwise specified, we perturb a sample with probability $0.9$ and use $t_{\min}{=}0.5$ for the first 350 epochs, then switch to $t_{\min}{=}0.7$ for the remaining 50 epochs. The same perturbed latent sequence is used by the pixel decoder to produce the decoded pixel inputs for the AR pass. We apply LayerNorm without affine parameters to intermediate states before diffusion-head supervision and after diffusion-head sampling.

\paragraph{Clean targets under encoder masking.}
When encoder masking is active, the implementation uses a clean no-mask encoder pass to define the detached target $\mathbf{z}_i$ for the AR diffusion loss, while the masked/noised path is used to train the decoder and construct decoded pixel inputs. The AR loss therefore does not backpropagate through the detached target-producing encoder path; the encoder and decoder are trained through reconstruction, perceptual, and auxiliary representation objectives.

\paragraph{Encoder-side gradient scaling.}
The reconstruction and autoregressive generation objectives share the same AR backbone. To balance the training signal strength, we scale the gradients flowing from the encoder-side reconstruction and auxiliary losses into the shared AR backbone by a constant factor. Unless otherwise specified, we use a gradient scale of $0.3$.

\paragraph{AR rectified-flow loss.}
For each token target $\mathbf{z}_i$, we draw $\epsilon_i\sim\mathcal{N}(0,I)$ and $s_i\sim\operatorname{LogitNormal}(0,1)$, where $\operatorname{LogitNormal}(0,1)$ denotes the distribution obtained by applying the sigmoid function to a standard Gaussian sample. We train the diffusion head on the linear interpolation $\mathbf{z}^{s_i}_i=(1-s_i)\epsilon_i+s_i\mathbf{z}_i$:
\begin{equation}
    \mathcal{L}_{\rm AR}
    = \frac{1}{T}\sum_{i=1}^{T}
    \left\|v_\omega(\mathbf{z}^{s_i}_i,s_i,\bar{\mathbf{h}}_{i-1})
    - (\mathbf{z}_i-\epsilon_i)\right\|_2^2,
\end{equation}
where $\bar{\mathbf{h}}_{i-1}$ is computed from the decoded prefix $\bar{\mathbf{x}}_{<i}$ rather than the clean prefix.

\paragraph{Reconstruction and auxiliary losses.}
The pixel decoder is supervised on the reconstructed image with an $\ell_1$ reconstruction loss and an LPIPS perceptual loss~\citep{zhang2018unreasonable}. We also use an auxiliary representation loss that predicts per-patch normalized pixel values from the AR hidden states produced in the reconstruction pass. The overall objective is
\begin{equation}
    \mathcal{L}
    = \mathcal{L}_{\rm AR}
    + \lambda_{\rm rec}\mathcal{L}_{\rm rec}
    + \lambda_{\rm lpips}\mathcal{L}_{\rm lpips}
    + \lambda_{\rm repr}\mathcal{L}_{\rm repr} .
\end{equation}
For the main runs, we use $\lambda_{\rm rec}{=}1$, $\lambda_{\rm lpips}{=}1$, and $\lambda_{\rm repr}{=}1$ unless otherwise stated in ablations.

\paragraph{Ground-truth pixel replacement.}
We replace a small fraction of decoded AR inputs with noised ground-truth pixel patches. This stabilizes training while preserving the decoded-input training distribution. We set the replacement probability to $0.04$ by default.

\section{Sampling Details}
\label{app:sampling_details}

At inference time, generation proceeds sequentially in raster order. At step $i$, the AR Transformer consumes the previously generated pixel patch $\hat{\mathbf{x}}_{i-1}$, produces a hidden state, the diffusion head samples $\hat{\mathbf{z}}_i$, and the pixel decoder maps $\hat{\mathbf{z}}_i$ to $\hat{\mathbf{x}}_i$. We use KV caching for both the AR Transformer and the causal pixel decoder.

For the velocity-parameterized diffusion head used in the main experiments, sampling starts from Gaussian noise at $s=0$. We take 100 Euler--Maruyama steps over $s\in[0,0.96]$, followed by one deterministic Euler step to $s=1$. We use classifier-free guidance by evaluating conditional and null-label predictions in the diffusion head, and apply a linear guidance schedule over autoregressive positions. As in training, the generated intermediate state is normalized before being passed to the pixel decoder.

\section{Training Hyperparameters}
\label{app:training_hparams}

All main ImageNet-1K models are trained for 400 epochs with global batch size 512. We use AdamW with peak learning rate $3\times10^{-4}$, $\beta_1{=}0.9$, $\beta_2{=}0.95$, weight decay $0.05$, gradient clipping at norm $1.0$, a cosine learning-rate schedule, and 20K warmup steps. We maintain exponential moving averages (EMA) of model weights with decay 0.9999. Unless otherwise specified, class dropout is $0.1$, the number of class prefix tokens is 16, the patch size is 16, and the intermediate dimension is $d_z{=}16$.


\clearpage
\section{Rollout Context Alignment}
\label{app:rollout_diagnostics}

PRA constructs training inputs through the same intermediate-state-to-pixel path used during inference. We examine whether the resulting AR contexts better approximate those encountered during sequential rollout, compared with a pixel-space AR baseline trained with direct noise injection into ground-truth pixels.

For each model, we construct its training inputs and generate 50K class-balanced samples by full sequential rollout. We then compare the distributions of AR hidden features induced by its training-input prefixes and by its own generated prefixes. We compute sliced Wasserstein-1 ($\mathrm{SW}_1$), sliced Wasserstein-2 ($\mathrm{SW}_2$), and hidden Fr\'{e}chet distance separately at each patch position, and report the average of each metric over all 256 positions. Each reported distance therefore measures a model's training--inference gap in the hidden representations that condition patch prediction.

\begin{table}[h]
    \centering
    \small
    \caption{Alignment between training and sequential-rollout contexts. Distances are computed in each model's AR hidden-feature space and averaged over 256 patch positions; lower is better.}
    \label{tab:rollout_alignment}
    \setlength{\tabcolsep}{8pt}
    \begin{tabular}{lccc}
        \toprule
        Model & $\mathrm{SW}_1\downarrow$ & $\mathrm{SW}_2\downarrow$ & Hidden Fr\'{e}chet$\downarrow$ \\
        \midrule
        Pixel-space AR + input noise & 0.26955 & 0.33659 & 170.918 \\
        PRA & \textbf{0.07681} & \textbf{0.09593} & \textbf{25.277} \\
        \bottomrule
    \end{tabular}
\end{table}

As shown in Table~\ref{tab:rollout_alignment}, PRA has smaller measured training--inference gaps under all three metrics. These results support PRA's intended approximation of inference-time contexts at the level of the AR representations used for patch prediction.

We also measure how far a rollout drifts from the corresponding training inputs in pixel space. For each model, we construct a sequence of training inputs, use its first 128 patches as the prefix, and generate the remaining 128 patches autoregressively. We measure the $L_1$ distance between each generated patch and the training input at the same position in that sequence. Table~\ref{tab:rollout_drift} shows that the distance increases along the generated suffix for both models, but stays lower for PRA in all four windows.

\begin{table}[h]
    \centering
    \small
    \caption{Pixel-space rollout drift. Columns give the mean $L_1$ distance to the corresponding training inputs over successive 32-patch windows (lower is better).}
    \label{tab:rollout_drift}
    \setlength{\tabcolsep}{6pt}
    \begin{tabular}{lcccc}
        \toprule
        Model & 129--160 & 161--192 & 193--224 & 225--256 \\
        \midrule
        Pixel-space AR + input noise & 0.398 & 0.480 & 0.509 & 0.515 \\
        PRA & \textbf{0.249} & \textbf{0.319} & \textbf{0.334} & \textbf{0.336} \\
        \bottomrule
    \end{tabular}
\end{table}

\clearpage
\section{Model Generated Examples}
\label{app:examples}

We show selected 256$\times$256 samples generated by \emph{PRA-L}, from
Fig.~\ref{fig:sample_start} to Fig.~\ref{fig:sample_end}. Each figure groups
examples from one ImageNet class and reports the class name and label.

\begingroup
\captionsetup{skip=5pt}

\begin{figure}[H]
    \centering
    \vbox{%
        \offinterlineskip
        \hbox{%
            \includegraphics[width=0.2\linewidth]{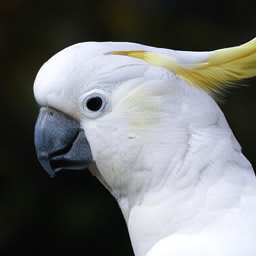}%
            \includegraphics[width=0.2\linewidth]{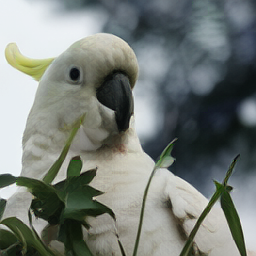}%
            \includegraphics[width=0.2\linewidth]{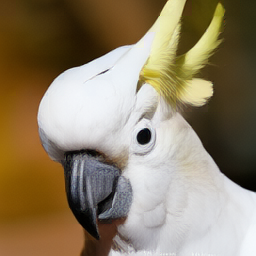}%
            \includegraphics[width=0.2\linewidth]{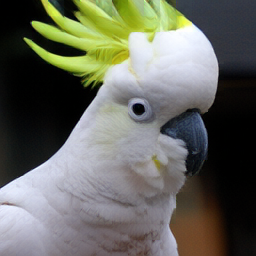}%
            \includegraphics[width=0.2\linewidth]{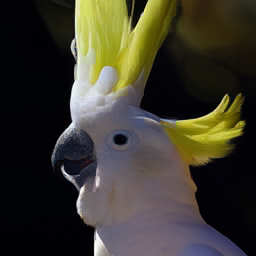}%
        }%
        \hbox{%
            \includegraphics[width=0.2\linewidth]{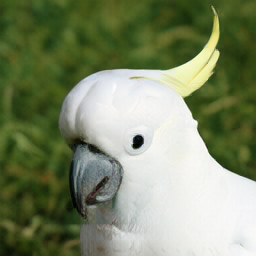}%
            \includegraphics[width=0.2\linewidth]{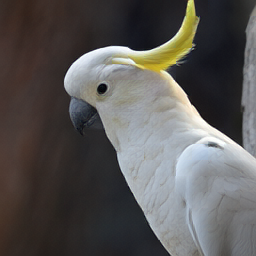}%
            \includegraphics[width=0.2\linewidth]{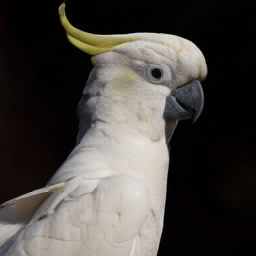}%
            \includegraphics[width=0.2\linewidth]{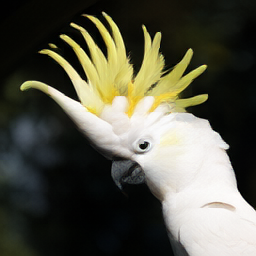}%
            \includegraphics[width=0.2\linewidth]{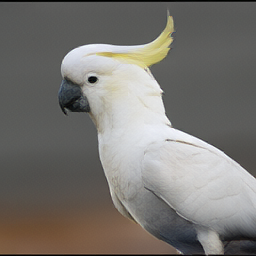}%
        }%
    }
    \caption{Selected 256$\times$256 \emph{PRA-L} samples. Class label: ``sulphur-crested cockatoo'' (89). \looseness=-1}
    \label{fig:samples_89}
    \label{fig:sample_start}
\end{figure}

\begin{figure}[H]
    \centering
    \vbox{%
        \offinterlineskip
        \hbox{%
            \includegraphics[width=0.2\linewidth]{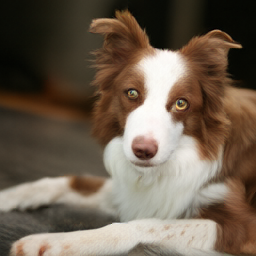}%
            \includegraphics[width=0.2\linewidth]{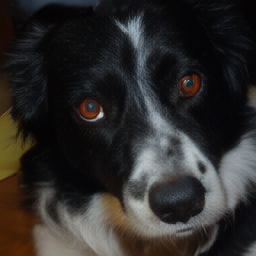}%
            \includegraphics[width=0.2\linewidth]{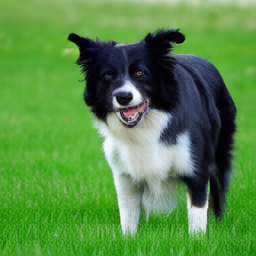}%
            \includegraphics[width=0.2\linewidth]{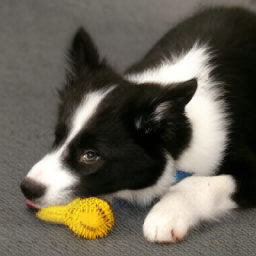}%
            \includegraphics[width=0.2\linewidth]{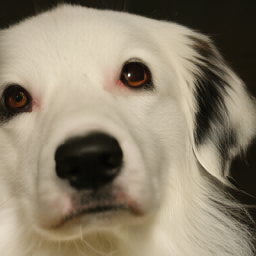}%
        }%
        \hbox{%
            \includegraphics[width=0.2\linewidth]{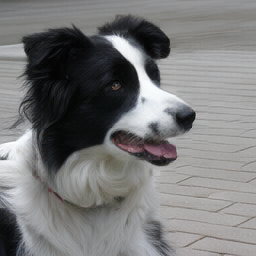}%
            \includegraphics[width=0.2\linewidth]{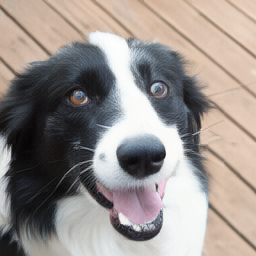}%
            \includegraphics[width=0.2\linewidth]{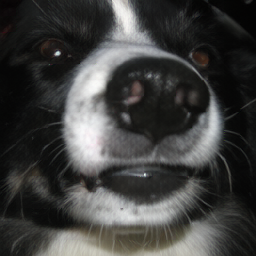}%
            \includegraphics[width=0.2\linewidth]{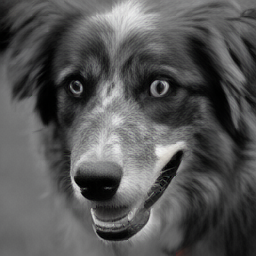}%
            \includegraphics[width=0.2\linewidth]{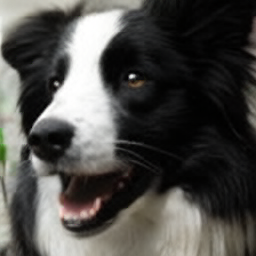}%
        }%
    }
    \caption{Selected 256$\times$256 \emph{PRA-L} samples. Class label: ``Border collie'' (232). \looseness=-1}
    \label{fig:samples_232}
\end{figure}

\begin{figure}[H]
    \centering
    \vbox{%
        \offinterlineskip
        \hbox{%
            \includegraphics[width=0.2\linewidth]{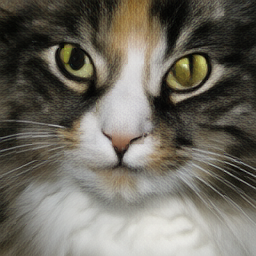}%
            \includegraphics[width=0.2\linewidth]{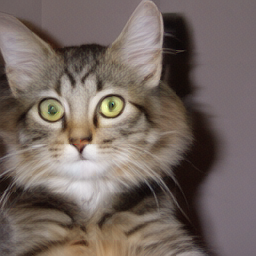}%
            \includegraphics[width=0.2\linewidth]{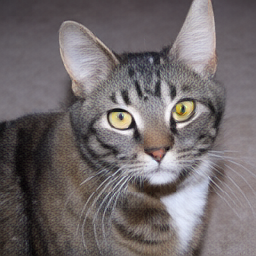}%
            \includegraphics[width=0.2\linewidth]{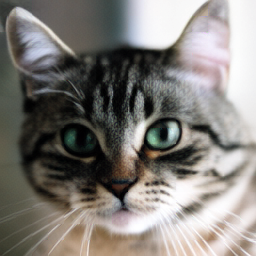}%
            \includegraphics[width=0.2\linewidth]{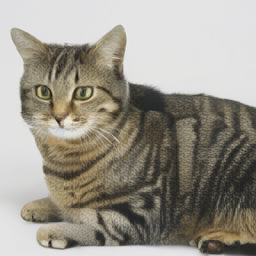}%
        }%
        \hbox{%
            \includegraphics[width=0.2\linewidth]{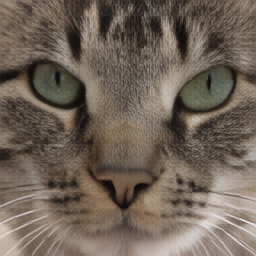}%
            \includegraphics[width=0.2\linewidth]{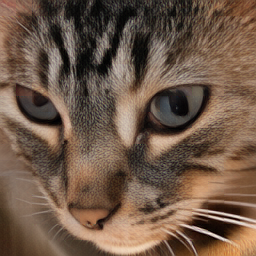}%
            \includegraphics[width=0.2\linewidth]{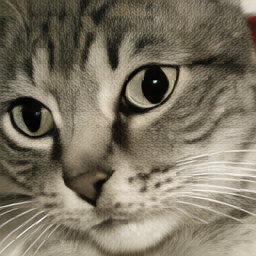}%
            \includegraphics[width=0.2\linewidth]{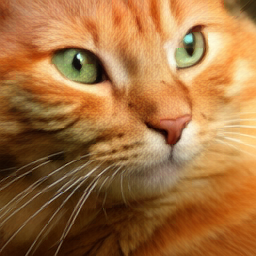}%
            \includegraphics[width=0.2\linewidth]{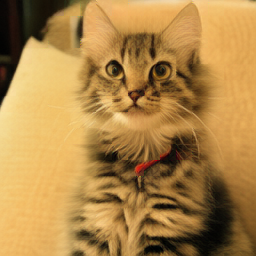}%
        }%
    }
    \caption{Selected 256$\times$256 \emph{PRA-L} samples. Class label: ``tabby cat'' (281). \looseness=-1}
    \label{fig:samples_281}
\end{figure}

\begin{figure}[H]
    \centering
    \vbox{%
        \offinterlineskip
        \hbox{%
            \includegraphics[width=0.2\linewidth]{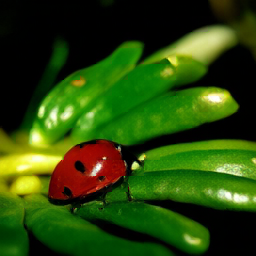}%
            \includegraphics[width=0.2\linewidth]{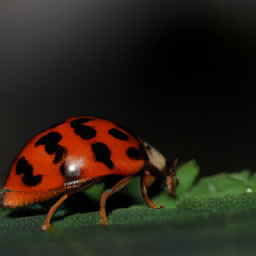}%
            \includegraphics[width=0.2\linewidth]{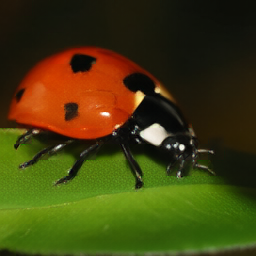}%
            \includegraphics[width=0.2\linewidth]{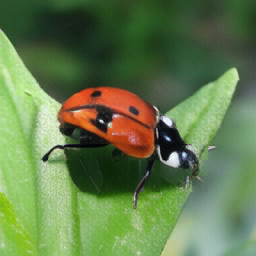}%
            \includegraphics[width=0.2\linewidth]{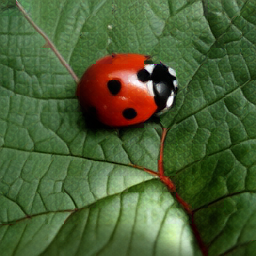}%
        }%
        \hbox{%
            \includegraphics[width=0.2\linewidth]{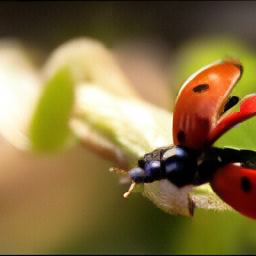}%
            \includegraphics[width=0.2\linewidth]{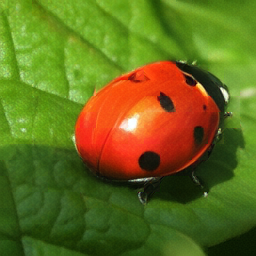}%
            \includegraphics[width=0.2\linewidth]{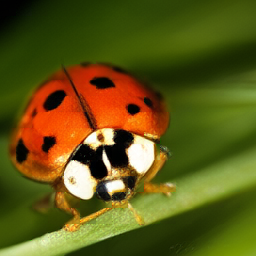}%
            \includegraphics[width=0.2\linewidth]{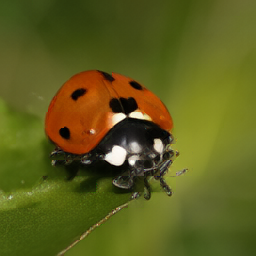}%
            \includegraphics[width=0.2\linewidth]{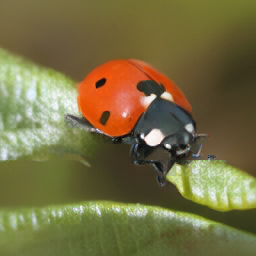}%
        }%
    }
    \caption{Selected 256$\times$256 \emph{PRA-L} samples. Class label: ``ladybug'' (301). \looseness=-1}
    \label{fig:samples_301}
\end{figure}

\begin{figure}[H]
    \centering
    \vbox{%
        \offinterlineskip
        \hbox{%
            \includegraphics[width=0.2\linewidth]{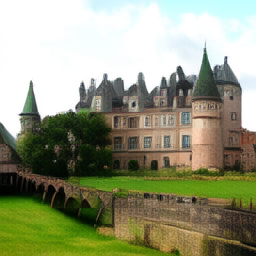}%
            \includegraphics[width=0.2\linewidth]{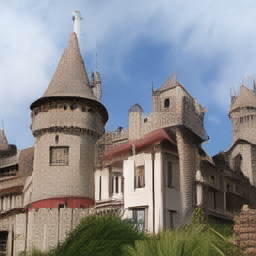}%
            \includegraphics[width=0.2\linewidth]{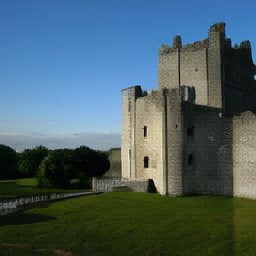}%
            \includegraphics[width=0.2\linewidth]{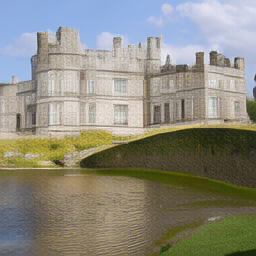}%
            \includegraphics[width=0.2\linewidth]{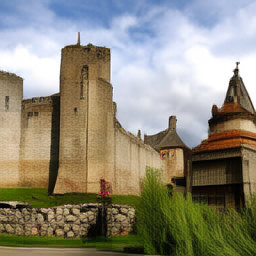}%
        }%
        \hbox{%
            \includegraphics[width=0.2\linewidth]{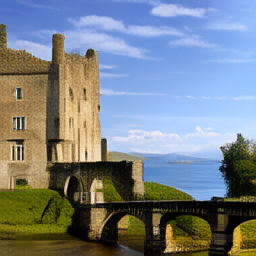}%
            \includegraphics[width=0.2\linewidth]{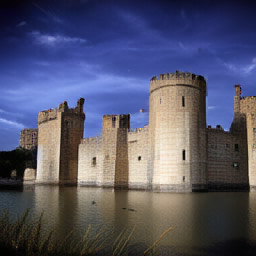}%
            \includegraphics[width=0.2\linewidth]{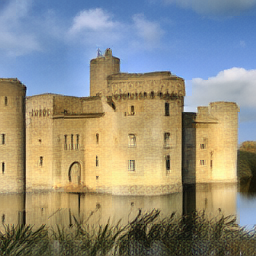}%
            \includegraphics[width=0.2\linewidth]{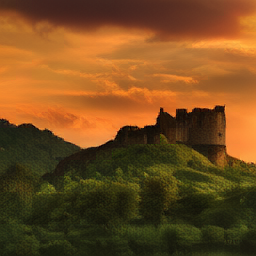}%
            \includegraphics[width=0.2\linewidth]{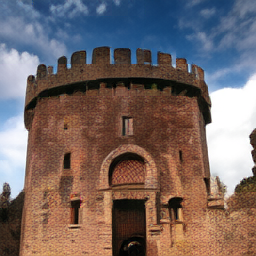}%
        }%
    }
    \caption{Selected 256$\times$256 \emph{PRA-L} samples. Class label: ``castle'' (483). \looseness=-1}
    \label{fig:samples_483}
\end{figure}

\clearpage

\begin{figure}[H]
    \centering
    \vbox{%
        \offinterlineskip
        \hbox{%
            \includegraphics[width=0.2\linewidth]{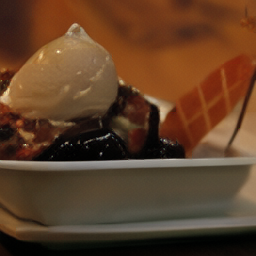}%
            \includegraphics[width=0.2\linewidth]{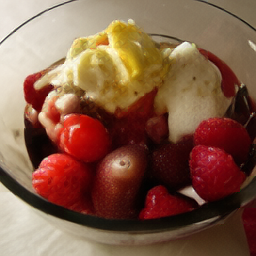}%
            \includegraphics[width=0.2\linewidth]{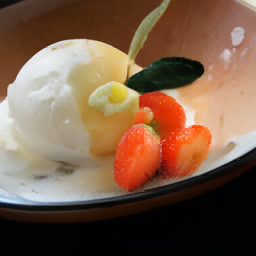}%
            \includegraphics[width=0.2\linewidth]{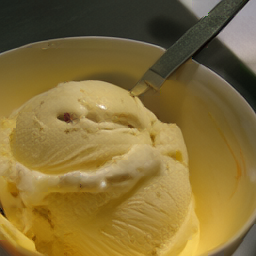}%
            \includegraphics[width=0.2\linewidth]{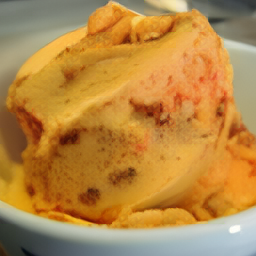}%
        }%
        \hbox{%
            \includegraphics[width=0.2\linewidth]{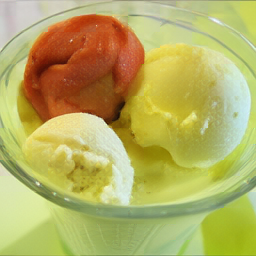}%
            \includegraphics[width=0.2\linewidth]{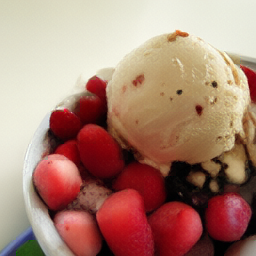}%
            \includegraphics[width=0.2\linewidth]{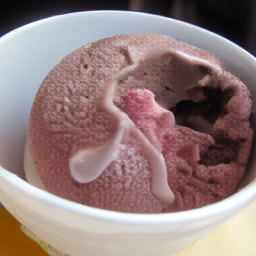}%
            \includegraphics[width=0.2\linewidth]{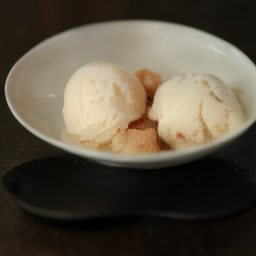}%
            \includegraphics[width=0.2\linewidth]{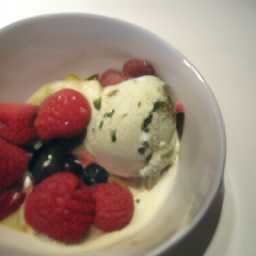}%
        }%
    }
    \caption{Selected 256$\times$256 \emph{PRA-L} samples. Class label: ``ice cream'' (928). \looseness=-1}
    \label{fig:samples_928}
\end{figure}

\begin{figure}[H]
    \centering
    \vbox{%
        \offinterlineskip
        \hbox{%
            \includegraphics[width=0.2\linewidth]{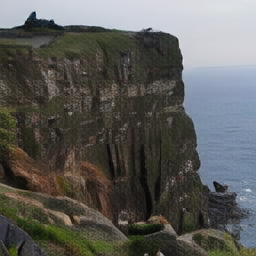}%
            \includegraphics[width=0.2\linewidth]{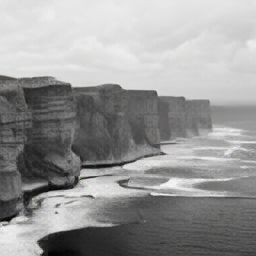}%
            \includegraphics[width=0.2\linewidth]{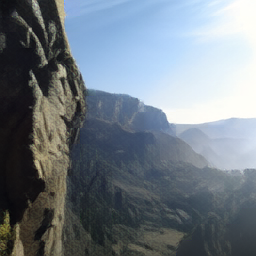}%
            \includegraphics[width=0.2\linewidth]{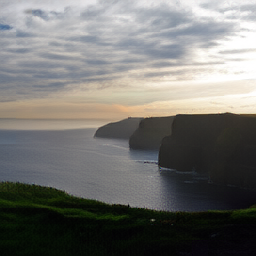}%
            \includegraphics[width=0.2\linewidth]{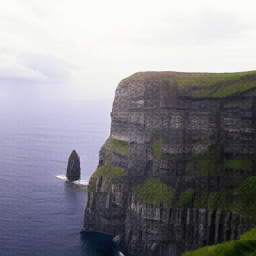}%
        }%
        \hbox{%
            \includegraphics[width=0.2\linewidth]{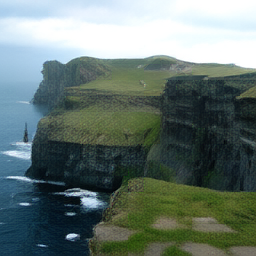}%
            \includegraphics[width=0.2\linewidth]{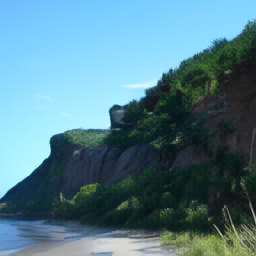}%
            \includegraphics[width=0.2\linewidth]{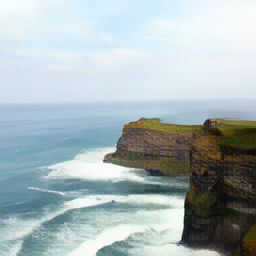}%
            \includegraphics[width=0.2\linewidth]{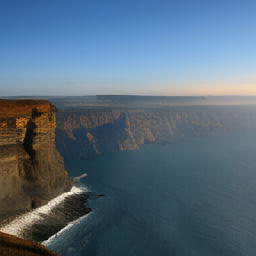}%
            \includegraphics[width=0.2\linewidth]{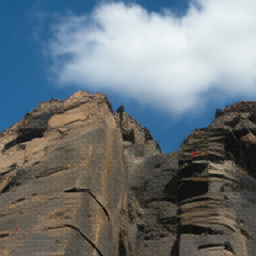}%
        }%
    }
    \caption{Selected 256$\times$256 \emph{PRA-L} samples. Class label: ``cliff'' (972). \looseness=-1}
    \label{fig:samples_972}
    \label{fig:sample_end}
\end{figure}

\endgroup

\end{document}